%% file: vita_slam_vitac_workshop.tex
\documentclass[letterpaper, 10 pt, journal, twoside]{ieeeconf}  % Comment this line out if you need a4paper
\IEEEoverridecommandlockouts                              % This command is only needed if you want to use the \thanks command
\usepackage{capt-of,graphicx,subcaption}
\usepackage{xargs}                      % Use more than one optional parameter in a new commands
\usepackage{hyperref}
\newcommand{\subparagraph}{}
\usepackage[compact]{titlesec}
\usepackage{amsmath,amssymb,bm,mathtools}
\usepackage{cite}
\usepackage[font={small,it}]{caption}
\usepackage{relsize}
\usepackage[utf8]{inputenc}
\usepackage[T1]{fontenc}
\usepackage{textcomp}
\usepackage{gensymb} %% for the degree symbol
\usepackage{tikz}
\usetikzlibrary{shadings}
\usepackage[margin=1.5cm]{geometry} %% Modify page margins if needed
\usepackage[subpreambles]{standalone} %% include standalone pseudocodes

\newcommand{\AlgName}{ViTa-SLAM}
\newcommand{\RobotName}{WhiskEye}

\graphicspath{{./sections/figures/}} % Define path to folder containing figures

\title{\LARGE \textbf{\AlgName: Biologically-Inspired Visuo-Tactile SLAM}}

\author{Oliver Struckmeier, Kshitij Tiwari, Martin J. Pearson, and Ville Kyrki
\thanks{This research has received funding from the European Union’s Horizon 2020 Framework Programme for Research and Innovation under the Specific Grant Agreement No. 785907 (Human Brain Project SGA2).}
\thanks{Oliver Struckmeier, Kshitij Tiwari and Ville Kyrki are with the Department of Electrical Engineering and Automation, Aalto University, Espoo 02150, Finland \tt \footnotesize \{first.last\}@aalto.fi}
\thanks{Martin J. Pearson is with the Bristol Robotics Laboratory, Bristol BS16 1QY, U.K (\tt\footnotesize martin.pearson@brl.ac.uk)}
}

\begin{document}
\maketitle
\input{sections/abstract}%% abstract

%% Introduction
\input{sections/intro}

%% Method
\input{sections/method}

%% Empirical Evaluation
\input{sections/evaluation}

%% Conclusion and Future works
\input{sections/conclusion}

%%% Bibliography
\bibliographystyle{ieeetr}
\bibliography{vita_slam_vitac_workshop}%

\end{document}

%% file: sections/abstract.tex
\begin{abstract}
In this work, we propose a novel, bio-inspired multi-sensory SLAM approach called \AlgName. Compared to other multi-sensory SLAM variants, this approach allows for a seamless multi-sensory information fusion whilst naturally interacting with the environment. The algorithm is empirically evaluated in a simulated setting using a biomimetic robot platform called the \RobotName. Our results show promising performance enhancements over existing bio-inspired SLAM approaches in terms of loop-closure detection.
\end{abstract}

%% file: sections/intro.tex
\section{Introduction}
Of late, there is growing interest in biologically inspired SLAM algorithms. For instance, the rat hippocampus inspired RatSLAM \cite{milford2004ratslam} provides a biologically-inspired $3$ DOF SLAM architecture which has been shown to work for robots operating under a myriad of environmental conditions. While the vanilla RatSLAM was purely designed as a visual SLAM approach, there are additional works that now explain its usage with other sensory modalities like WiFi \cite{berkvens2014biologically} and auditory signals \cite{steckel2013batslam}. 

\begin{figure}[!htbp]
\centering
\includegraphics[scale=0.15]{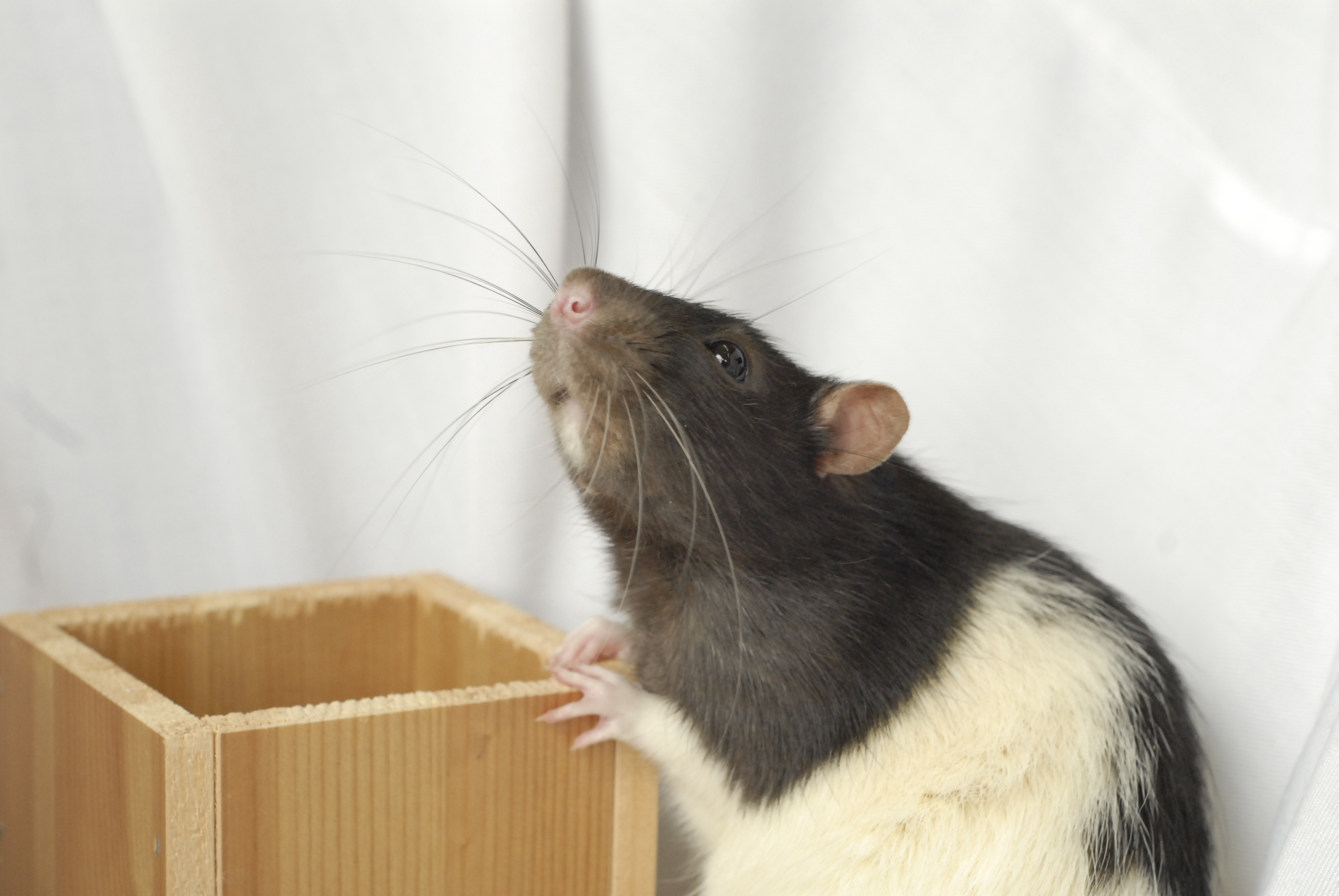}
\caption{Rat interacting with its surroundings while foraging.}
\label{fig:blind_rat}
\end{figure}

In nature, most mammals like the rats use their whiskers to interact with their environment through contact (Fig.~\ref{fig:blind_rat}). In \cite{cheung2012maintaining}, it was shown that rats rely on whiskers to maintain a cognitive understanding of their surroundings. To this end, this work describes our preliminary findings of extending the RatSLAM algorithm to account for both visual and tactile sensory modalities. The new algorithm will be hereby referred to as \AlgName. The novel aspect of this algorithm is that it utilizes multiple sensory modalities to estimate the state of the robot which are implicitly fused as opposed to other methods that require explicit weighted fusion like \cite{milford2013brain}.

%% file: sections/method.tex
\section{\AlgName}
\AlgName~comprises of two sensory inputs: \textit{visual} and \textit{tactile}. Below, we elucidate how each of these sensory modalities are processed to obtain the best pose estimate.

\subsection{Perceputal Data Preprocessing}
Each visual frame is converted to greyscale (B/W img) and processed to obtain a \textbf{local view template} $(V)$ similar to the vanilla RatSLAM \cite{milford2004ratslam}.

The tactile data contains two kinds of information: \textit{contact points} and \textit{deflection}. \textit{Contact points} $(Cts.)$ refers to the 3D point on the object surface in the world frame where the contact is made. These points are obtained by transforming the whisker contact points from a head centric frame to the world frame and are used to obtain a Point Feature Histogram (PFH)~\cite{rusu2008learning}.

The \textit{deflection} $(Defl.)$ refers to the amount of bending of the whiskers and is used to obtain the Slope Distribution Array (SDA)~\cite{kim2007biomimetic}. Together, they are referred to as \textbf{tactile features} $(T)$. Previous attempts at pure whisker sensor based SLAM can be found in works like the WhiskerRatSLAM \cite{salman2018whisker}.

\subsection{Overall System Architecture}
\begin{figure}[!htbp]
\centering
\resizebox{.8\linewidth}{!}{
\input{sections/figures/vita_slam}}
\caption{Overview of Vita-SLAM architecture.}
\label{fig:vitaslam_architecture}
\end{figure}
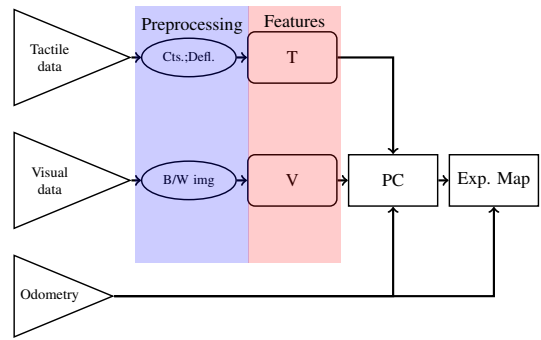

The overall system architecture is shown in Fig.~\ref{fig:vitaslam_architecture} wherein the pose cell network (PC)~\cite{ball2013openratslam} combines the information $V,T$ from both modalities to obtain the best pose estimate. A semi-metric experience map (Exp. Map)~\cite{milford2005experience} is generated to evaluate the model performance.

%% file: sections/figures/vita_slam.tex
\begin{tikzpicture}[
				round boxes/.style={draw, rectangle,%
                thick,minimum height=1cm, rounded corners,
                minimum width=1cm, black, text=black,
                text width=15mm, anchor=center, align=center},
                boxes/.style={draw, rectangle,%
                thick,minimum height=1cm,
                minimum width=1cm, black, text=black,
                text width=15mm, anchor=center, align=center}, 
    			% Define styles for some special nodes
   				right iso/.style={isosceles triangle,scale=0.8,sharp corners, anchor=center, xshift=-4mm},
    			left iso/.style={right iso, rotate=180, xshift=-8mm},
    			txt/.style={text width=1.5cm,anchor=center},
    			ellip/.style={ellipse,scale=0.75},
    			empty/.style={draw=none}
    			]
  \matrix (mat) [matrix of nodes, nodes=boxes, column sep=0.2cm, row sep=0.5cm] 
  {
                 &   &   &           &           \\ 
    |[right iso]|{Tactile data} & |[ellip]|{Cts.;Defl.} & |[round boxes]|{T} &           &           \\
    |[right iso]|{Visual data} & |[ellip]|{B/W img} & |[round boxes]|{V} & PC & Exp. Map \\
    |[right iso]|{Odometry}     &   &   &           &           \\
  };  
  
%% Node ordering: [row- column]
%% Tactile data
\draw [very thick, black, ->](mat-2-1)--(mat-2-2);
\draw [very thick, black, ->](mat-2-2)--(mat-2-3);
\draw [very thick, black, ->](mat-2-3)-|(mat-3-4);
%%
%%
%%%%% Visual data

\draw [very thick, black, ->](mat-3-1)--(mat-3-2);
\draw [very thick, black, ->](mat-3-2)--(mat-3-3);
\draw [very thick, black, ->](mat-3-3)--(mat-3-4);
\draw [very thick, black, ->](mat-3-4)--(mat-3-5);

%%
%%%% Odometry info connections
\draw [very thick, black, ->](mat-4-1)-|(mat-3-4);
\draw [very thick, black, ->](mat-4-1)-|(mat-3-5);

%%
%%%% Obj Map and Exp Map fusion to Vita Map
%\draw [very thick, black, ->](mat-3-5)-|(mat-4-6);
%\draw [very thick, black, ->](mat-5-5)-|(mat-4-6);

%% Bounding rectangles
% draw the rectangles
\node at(-2.75cm, .5cm) [right,fill=blue,text opacity=1,opacity=.2, minimum width=1.8cm, minimum height=5cm, text height=-4.cm] {Preprocessing};

\node at(-0.55cm, .5cm) [right,fill=red,text opacity=1,opacity=.2, minimum width=1.8cm, minimum height=5cm, text height=-4.2cm] {Features};

\end{tikzpicture}

%% file: sections/evaluation.tex
\section{Evaluation}
In this section, we describe the robot platform and the experimental scenario under which the model performance of \AlgName~was evaluated. Following suit, the results obtained are described.

\subsection{Robot Platform}
\begin{figure}[!htbp]
\centering
\includegraphics[clip=true, trim= {3.5cm 0 3.5cm 0},scale=0.1]{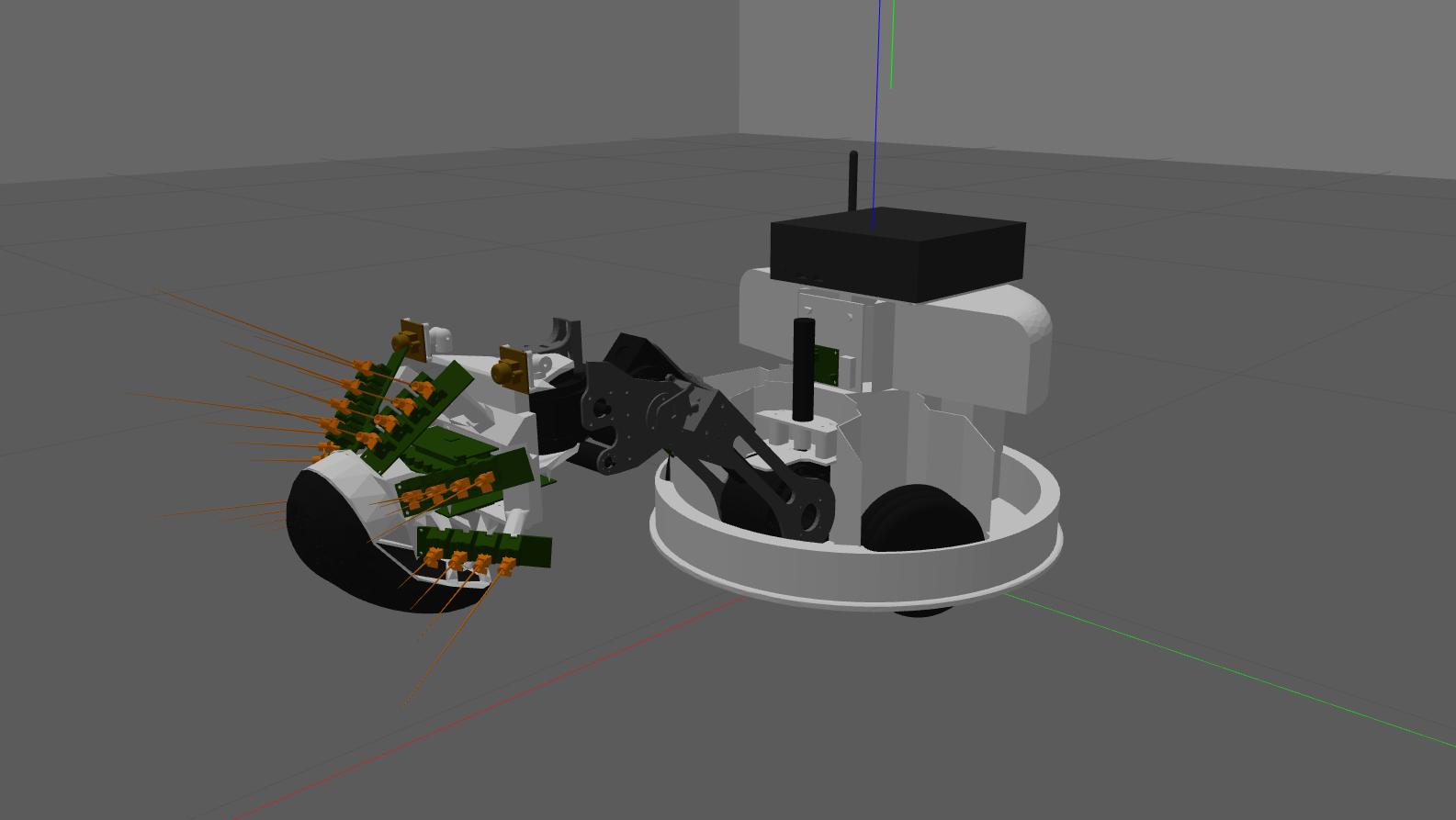}
\caption{Simulated robot platform, WhiskEye.}
\label{fig:whiskeye_sim}
\end{figure}

The robot platform used for empirical evaluation is called the \RobotName~and is shown in Fig.~\ref{fig:whiskeye_sim}. This robot is equipped with $24$ whiskers which can be individually controlled and comes with a mobile base. The analog data from the whiskers is sampled at $500$Hz. The tactile templates generated from the $3$D whisker contact points are published once per whisk cycle. The whiskers are mounted at the end of a neck which was kept fixed at a desired pose for this work. Aside from these, the platform is also equipped with a static HD ($1280 \times 720$ pixels) RGB camera.

\subsection{Environment}
The robot is deployed in a visually sparse scene meaning there are not many diverse visual cues to be processed. Additionally, to break the 1-fold rotational symmetry of the rectangular environment being used, two static landmarks (cylinder and cube) are also placed in the scene. 

\begin{figure}[!htbp]
\centering
\includegraphics[scale=0.1]{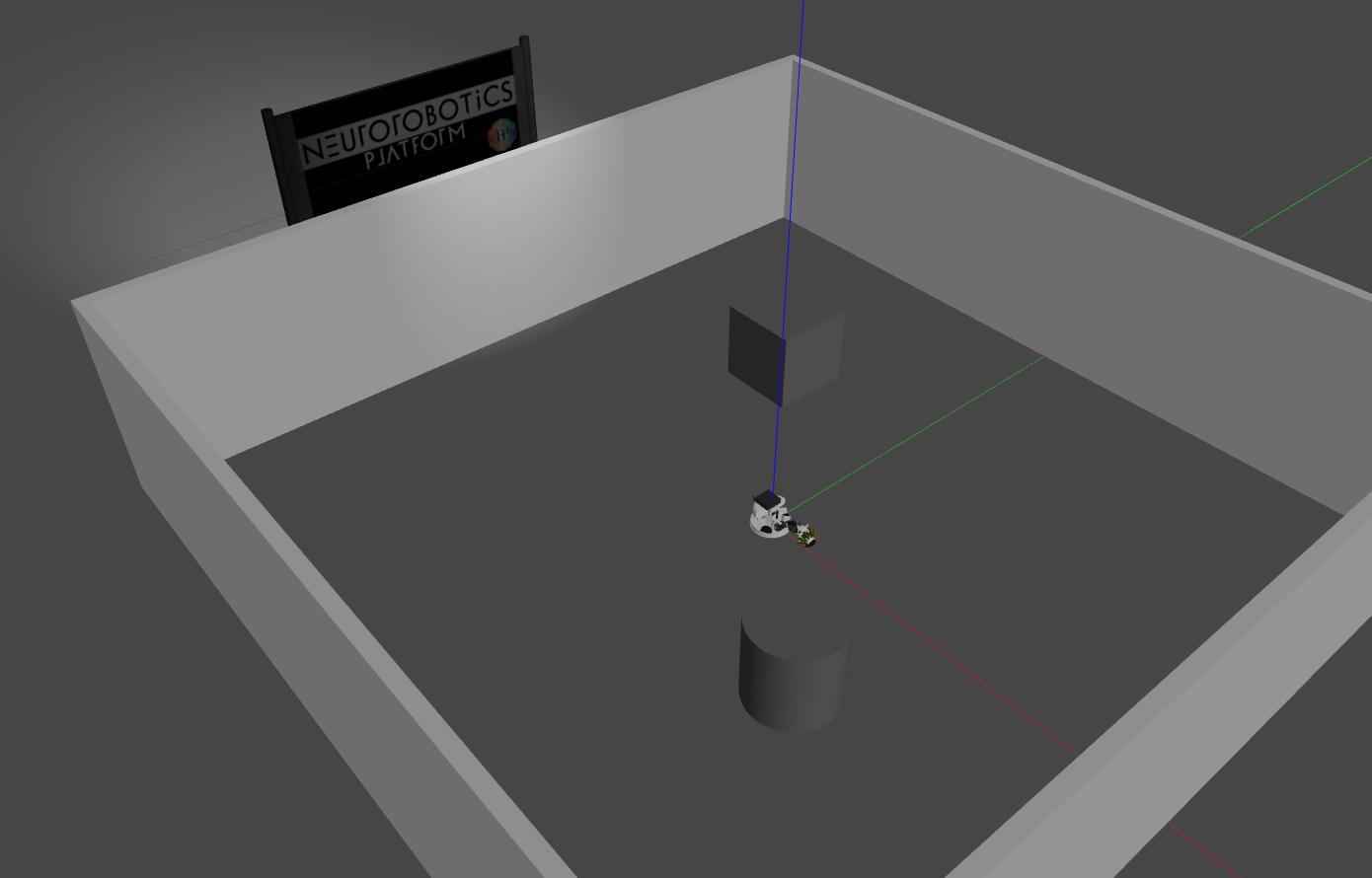}
\caption{Environmental setup. $2$ landmarks in a visually sparse arena.}
\label{fig:env}
\end{figure}

\subsection{Behavior}
The robot was given a pre-meditated trajectory to perform: It was required to reach landmark $1$ (cylinder), revolve around it, then approach landmark $2$ (cube), revolve around it and terminate exploration. At all times, the whiskers were being controlled using the Rapid Cessation of Protraction (RCP) protocol \cite{grant2009active}.

\subsection{Results}
\begin{figure}[!htbp]
\centering
\begin{minipage}{.5\linewidth}
  \centering
  \includegraphics[scale=0.13]{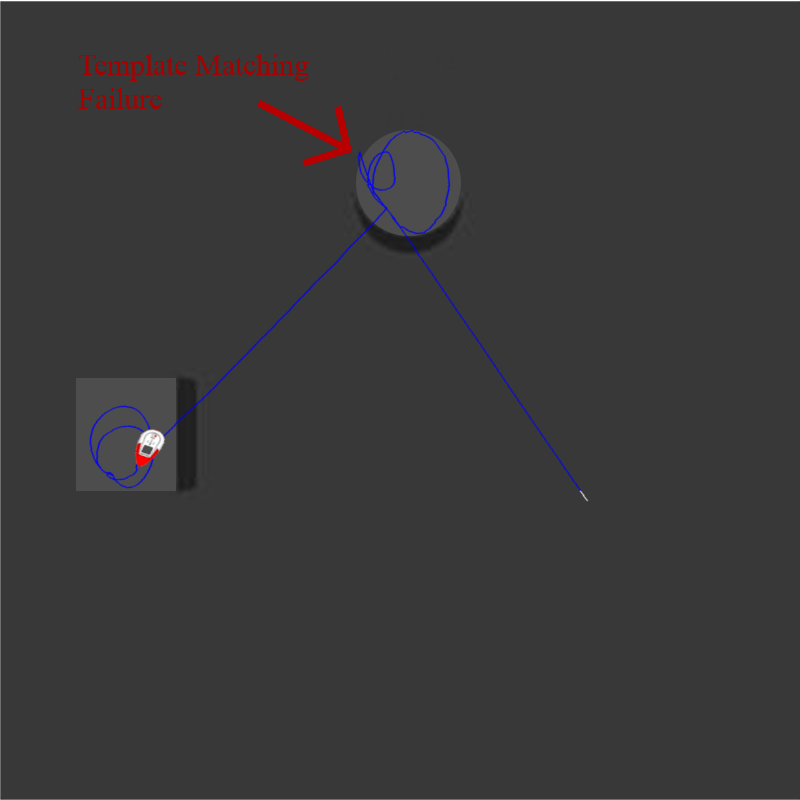}
  \captionof{figure}{RatSLAM failure.}
  \label{fig:ratslam_fail}
\end{minipage}%
\begin{minipage}{.5\linewidth}
  \centering
  \includegraphics[scale=0.13]{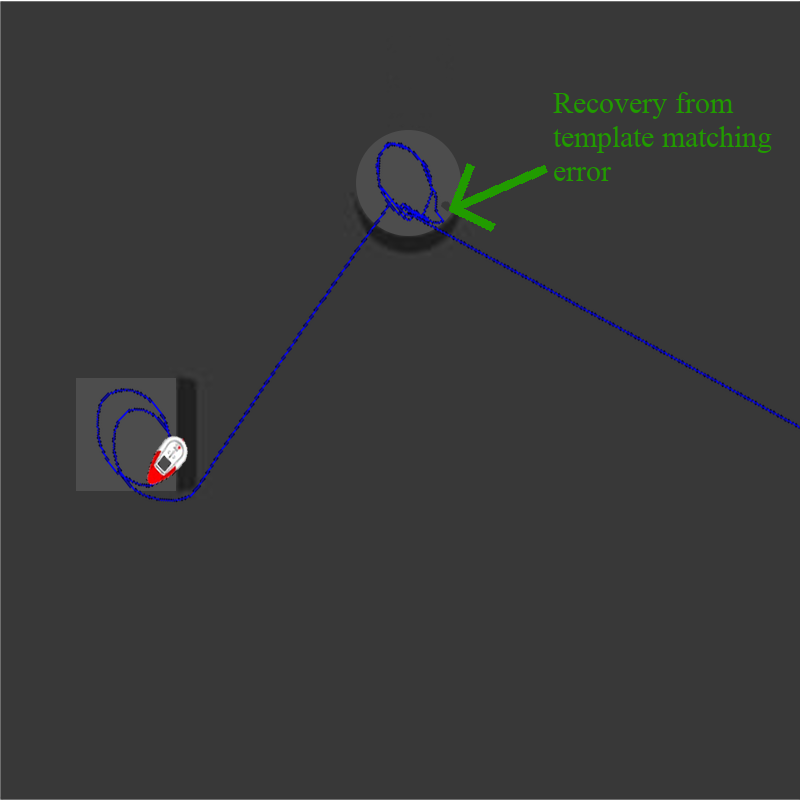}
  \captionof{figure}{\AlgName~success.}
  \label{fig:vita_slam_success}
\end{minipage}
\end{figure}

In order to evaluate \AlgName~performance, it was directly pitted against vanilla RatSLAM while keeping the platform and environment identical for both settings. As can be seen from Fig.~\ref{fig:ratslam_fail}, vanilla RatSLAM generated too many novel visual templates which eventually lead to failure of loop-closure detection. The frequent generation of novel templates can be attributed to visual sparsity of the scene. However, with \AlgName, additional tactile information helps detect loop closures\footnote{Video demonstration available \href{https://youtu.be/ygX340vA3rM}{here}.} as shown in Fig.~\ref{fig:vita_slam_success}. There are however challenges with using the passive whisking behavior which leads the whisker array of the robot to collide with landmarks. This induces noise into the tactile data which adversely affects loop-closure detection.

%% file: sections/conclusion.tex
\section{Conclusion and future works}

In this work, we presented preliminary variant of our novel \AlgName~ algorithm which allows for multi-sensory SLAM whilst interacting with the environment through contact.  Whilst the state-of-the-art bio-inspired SLAM model called RatSLAM was inspired by the rat hippocampal formations, it was designed purely for non-contact sensing scenarios. Similarly, WhiskerRatSLAM was designed purely for contact-sensing based SLAM. With this work, we have extended the outreach of these bio-inspired SLAM approaches to biomimetic robots bringing us one step closer to transitioning from biologically-inspired to biologically plausible methodologies.

In the future works, we plan on extending our algorithm to higher dimensions to account for the full $6$ DOF pose while the algorithm currently can handle upto $3$ DOF pose. While this poses significant computational challenges, it is essential to generalizing the applicability of this method. Additionally, we will investigate active sensory switch mechanism to minimize rudimentary sensory data acquisition. 